\documentclass[runningheads]{llncs}

\usepackage[normalem]{ulem}
\usepackage[greek,english]{babel}
\usepackage{amsmath}
\usepackage{amsfonts}

\usepackage{flushend}
\usepackage{xcolor}
\usepackage[pdftex]{graphicx}
\usepackage{float}
\usepackage{caption, subcaption}
\usepackage[export]{adjustbox}
\usepackage{wrapfig}

\usepackage{amssymb}
\usepackage{pifont}
\newcommand{\cmark}{\ding{51}}%
\newcommand{\xmark}{\ding{55}}%
\usepackage{multirow}
\usepackage{multicol}
\usepackage{arydshln}
\usepackage[edges]{forest}
\usepackage{nccmath}
\usepackage{caption}

\newcommand{\eat}[1]{}

\usepackage[official]{eurosym}
\usepackage{enumitem}
\setlist{nolistsep,noitemsep}
\usepackage{hyperref}

\begin{document}

\title{Temporal Sepsis Modeling: a Relational and Explainable-by-Design Framework}

\titlerunning{Temporal Sepsis Modeling}

\author{Vincent Lemaire\inst{1} \and
Nédra Mellouli\inst{2} \and
Pierre Jaquet\inst{3} 
 }

\authorrunning{}

\institute{
Orange Research, France, \email{vincent.lemaire@orange.com}
\and De Vinci Research Center, LIASD-UP8, France, \email{nedra.mellouli@devinci.fr}
\and Recherche Clinique La Fontaine, France, \email{pierre.jaquet@ch-stdenis.fr}
}

\maketitle

\begin{abstract}
Sepsis remains one of the most complex and heterogeneous syndromes 
in intensive care. While deep learning models achieve competitive performance in early sepsis prediction, their decision processes often remain difficult to interpret clinically, and explainability is typically added only through post-hoc methods. We propose an 
explainable-by-design framework based on a relational approach: 
temporal EHR data are represented in a relational schema, flattened 
via MDL-based propositionalisation into compact human-readable 
features, and classified using a selective Fractional Naive Bayes 
classifier. Evaluated on MIMIC-III (3,940 patients, 10-fold 
cross-validation), our approach achieves AUC = 0.983 - competitive 
with XGBoost (0.985) and CatBoost (0.985), and superior to LSTM 
(0.945)  - with only 98 selected variables and a 1 MB model footprint. 
Unlike post-hoc methods, interpretability is native and fourfold: 
univariate, global, local, and counterfactual.
\end{abstract}

\section{Introduction}

Sepsis \cite{OBRIEN20071012} is defined as a life-threatening organ dysfunction caused by an aberrant dysregulated host response to infection, and progression to septic shock markedly increases the risk of multi-organ failure and death. Early identification and risk stratification using validated clinical criteria, such as increases in SOFA score, enable rapid implementation of evidence-based interventions, including prompt broad-spectrum antimicrobial therapy and targeted resuscitation, which are associated with improved outcomes. Despite these advances, sepsis remains a leading cause of morbidity and mortality in intensive care worldwide.

Despite decades of clinical and computational research, early detection remains a major challenge due to the disease’s heterogeneity and temporal complexity. Traditional machine learning models treat sepsis as a binary event, neglecting the underlying patient-specific physiological patterns that drive its onset and progression. 
However, early detection of sepsis using machine learning algorithms remains an area of research that is still in an ongoing stage of development. Despite recent advances, existing models exhibit limited performance, with some failing to identify up to 67\% of sepsis cases \cite{Wong2021ExternalValidation}, which severely restricts their deployment in real-world clinical settings. To ensure that these tools can be used reliably across diverse patient populations, further studies focusing on their clinical implementation and evaluation under real-world conditions are required \cite{Fleuren2020Meta}.
Moreover, validating the reproducibility and generalization capabilities of these models remains a major challenge \cite{Ramspek2021ExternalValidation}. This difficulty is attributable to the fact that most existing studies have mainly focused on populations from intensive care units (ICUs) \cite{Shimabukuro2017SepsisPrediction}, thereby limiting the applicability of these models to other hospital departments. In addition, several retrospective studies have demonstrated that the success of machine learning models for sepsis detection depends heavily on the integration of clinically relevant and informative variables \cite{Fleuren2020Meta}.
Nevertheless, the development and evaluation of predictive models leveraging data from hospitalized patients across all clinical services remain limited. In this context, the MIMIC database provides a broad panel of clinical, biological, and physiological parameters, making it particularly well-suited for studying and comparing early sepsis prediction approaches. Accordingly, MIMIC-III \cite{johnson2016mimic} has become a widely used benchmark dataset in the literature for the comparative evaluation of sepsis prediction models and constitutes the experimental framework adopted in this study.

This paper is organized as follows. Section \ref{sec:context} reviews the main families of sepsis modeling. Section \ref{sec:relational} details the relational representation of patient trajectories, the technique of propositionalisation by flattening, and variable selection, emphasizing the ability of this method to produce understandable characteristics.
Section \ref{sec:results} then presents the models' performances, the quality of the aggregated variables, and the analysis of their univariate and local importance, while introducing 
counterfactual reasoning to explain and influence the model's decisions.
Finally, section \ref{sec:conclusion} summarizes the contributions of the proposed approach, highlighting its advantages in terms of explainability and flexibility.

\section{Context and Related Works}
\label{sec:context}



Early sepsis prediction from Electronic Health Records (EHRs) has attracted considerable attention in recent years, mainly due to the availability of high-frequency physiological measurements and the clinical importance of early intervention. Existing approaches differ primarily in how they represent temporal information and how they address model interpretability.

\subsection{Time-series and deep learning approaches}

Most recent works formulate early sepsis prediction as a multivariate time-series classification problem. Recurrent neural networks, particularly Long Short-Term Memory (LSTM) architectures, have been widely adopted to capture temporal dependencies in physiological signals \cite{futoma2017}. Extensions based on Temporal Convolutional Networks (TCNs) and attention mechanisms further improve performance by modeling long-range dependencies \cite{moor2020}. While these deep learning approaches often achieve state-of-the-art predictive accuracy, they require complex architectures and large amounts of data, and their decision processes remain difficult to interpret in clinical settings.

\subsection{Feature-based and window-aggregation approaches}
An alternative research approach involves transforming time-series data into fixed-length feature vectors by aggregating measurements over predefined temporal windows. This strategy enables the use of classical machine learning models such as logistic regression, random forests, or gradient boosting \cite{desautels}. Recently, Thiboud et al. \cite{thiboud2025development} proposed a gradient boosting model using aggregated clinical variables extracted from French hospital EHR data, showing competitive performance for early sepsis detection while preserving a degree of interpretability. However, most aggregation schemes are manually designed and do not explicitly control redundancy or model complexity.

\subsection{MDL-based and Bayesian explainable models}
While relational feature extraction provides a systematic way to transform patient trajectories into a flat attribute-value table suitable for classical machine learning, the large number of potential aggregate features can easily lead to overfitting and redundant representations. To address these limitations, Minimum Description Length (MDL)–based methods offer a principled framework for both constructing and selecting informative features. By explicitly balancing model fit and complexity, MDL-based approaches generate compact, high-quality aggregates that capture temporal patterns while remaining suitable for interpretable probabilistic classifiers \cite{Boulle2018}.
Explainability is further enhanced by selective Bayesian classifiers such as Fractional Naive Bayes \cite{hue2024fractionalnaivebayesfnb}, which aim to retain only informative and weakly dependent variables. In contrast to post-hoc explainability techniques applied to black-box models (e.g., SHAP \cite{lundberg2017}), these approaches provide intrinsic interpretability through probabilistic reasoning and explicit variable selection.

\subsection{Prediction horizons in early sepsis modeling}
The prediction horizon, denoted $h$, defines the lead time before sepsis onset at which a model aims to anticipate the condition. In classical machine learning approaches based on aggregated features \cite{desautels,thiboud2025development}, $h$ is typically fixed by the aggregation window: only measurements up to $h$ hours before onset are included in feature computation. This constrains the models to rely on recent observations and may limit long-range predictive power.
Deep learning models such as LSTM, TCN, and MGP-RNN
\cite{moor2020}
inherently handle sequences of arbitrary length and can be trained to predict multiple horizons simultaneously. For instance, models can be trained to output risk scores at $h = 1, 3, 6$ hours before sepsis onset, allowing a flexible trade-off between early detection and predictive accuracy. Multi-horizon prediction is particularly advantageous in ICU settings, where early warnings must be balanced against false alarms.

For relational and MDL-based  methods 
\cite{hue2024fractionalnaivebayesfnb}, the prediction horizon is explicitly controlled through the number of temporal observations $p$ included in the secondary table. Each patient trajectory is flattened over $p$ time steps corresponding to the desired horizon $h$, enabling consistent learning of aggregate variables that encode temporal dynamics relevant for early prediction. This approach allows the use of interpretable probabilistic classifiers while maintaining horizon-specific information. Overall, while deep learning methods naturally support multi-horizon prediction, feature-based and relational methods achieve horizon flexibility by careful window selection and aggregation, highlighting a key design choice in early sepsis modeling.

To address the interpretability limitations of existing methods and the lack of principled complexity control, we represent patient trajectories as a root table linked to a secondary table of temporal observations. MDL-based aggregation, combined with supervised discretization and selective Bayesian classification, yields compact, interpretable features suitable for varying prediction horizons.
Unlike post-hoc explainability methods applied to black-box models,  our approach provides intrinsic interpretability through probabilistic reasoning, without sacrificing predictive performance.
Our methodology details (i) relational data representation, (ii) feature extraction and flattening, and (iii) MDL-driven variable selection and Bayesian classification.

\section{Modeling: From temporal Data to Flat Data}
\label{sec:relational}

\subsection{Dataset}
\label{sec:protocol}





The MIMIC-III database \cite{johnson2016mimic} is a publicly available critical care dataset containing de-identified electronic health records of over 40,000 adult patients admitted to intensive care units at the Beth Israel Deaconess Medical Center between 2001 and 2012.
Each patient trajectory comprises 36 key physiological parameters, including vital signs (heart rate, blood pressure, respiratory rate, oxygen saturation), laboratory measurements (e.g., white blood cell count, creatinine), and administered interventions relevant to sepsis prediction. As previous studies on this dataset \cite{johnson2016mimic}, two variables are redrawn: `Hour' and `Gender' (the last one to prevent biased results). Measurements are primarily recorded hourly, although the sampling frequency varies across variables and patients. As is common in real-world electronic health records, missing values occur due to skipped measurements, delayed lab results, or clinical decisions not to collect certain tests. 
Variables with more than 20\% missing observations were excluded, in
line with standard practice in MIMIC-III-based studies~\cite{Fleuren2020Meta,moor2021sepsisreview}.
This threshold reflects a clinically motivated trade-off: variables
missing in more than one fifth of measurements are unlikely to provide
reliable signal for temporal modeling. For the remaining variables,
missing values were imputed using a nearest-neighbor interpolation
strategy applied independently per patient trajectory, preserving
temporal ordering.

A 12-hour observation window is clinically justified, as early hemodynamic and inflammatory dysregulations preceding sepsis onset\cite{moor2021sepsisreview}. We retained patients with 6 measurements before the onset-based index time and 6 measurements after that time for sepsis-positive trajectories. (see Figure \ref{fig:schema}) . Then the extracted dataset contains 3940 patients that will be used below in the experiments. Importantly, the resulting
cohort is approximately balanced between sepsis-positive and
sepsis-negative cases (roughly 50\% each), which ensures that standard
accuracy metrics are meaningful and that no class-rebalancing
procedure is required. While this selection reduces dataset size, it
guarantees label reliability and temporal completeness, two properties
that are essential for the validity of the relational modeling approach.

The choice of prediction horizons $h$ equal to $3$ and $6$ hours before sepsis onset is motivated by both clinical and technical considerations. Clinically, early intervention in sepsis is critical, as each hour of delayed treatment significantly increases mortality \cite{rudd2020global}. A 3-hour horizon provides a realistic window for timely clinical action, while a 6-hour horizon offers earlier warning to plan interventions and allocate resources. From a modeling perspective, MIMIC-III provides hourly measurements that allow the extraction of temporal patterns over 3- to 6-hour windows, which are sufficiently informative for sequential models such as LSTM, TCN, or MDL-based relational feature extraction combined with Bayesian classifiers.
\vspace{-4mm}

\subsection{Viewing the data as relational data}

\begin{figure}
\centering
\begin{minipage}{.48\textwidth}
  \centering
  \includegraphics[width=1.0\linewidth]{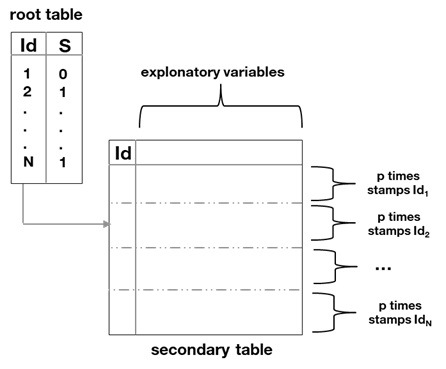}
  \captionof{figure}{A Time series relational data flattening}
  \label{fig:schema}
\end{minipage}
\hfill
\begin{minipage}{.48\textwidth}
  \centering
  \includegraphics[width=1.0\linewidth]{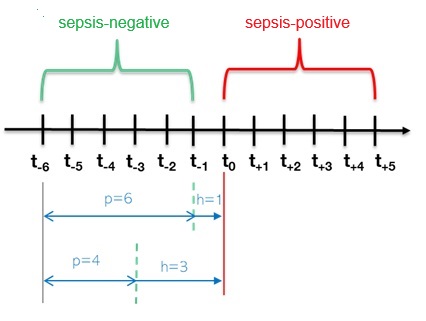}
  \captionof{figure}{Temporal relationship between the fixed observation window ($p=4,6$ hours) and different prediction horizons $h=3,1$.}
  \label{fig:ts}
\end{minipage}
\end{figure}

The dataset described in the previous section may be viewed as a relational dataset (see Figure \ref{fig:schema}) with a star schema. In this paradigm, the root table contains two columns: the first one is the `Id' of the patient (a reference to the patient), and the second one contains the class to predict (sepsis-positive or sepsis-negative). In the secondary table, the number of columns corresponds to the explanatory variables $d$ ($d=34$ excluding 'Hour' and 'Gender' variables) described in the previous section, plus the column `Id' to enable table joins. The number of lines of the secondary table is  $p  N$,  where $p$ is the number of (logs or time stamps) of each patient used to do the prediction, which depends on the horizon for which one would like the prediction, as described just below.

This is represented in the Figure \ref{fig:ts}. In the initial dataset, each patient is represented by a measure (a vector) of 34 variables, each one measured 12 times ($t \in \{t_{-6},...,t_{+5}\}$) every hour, so at different time stamps. In the prediction setting considered here, $t=t_0$ denotes the sepsis onset time for sepsis-positive patients. Therefore, to predict whether a patient trajectory is sepsis-positive or sepsis-negative one hour before $t=t_0$, one may use at most 6 measurements ($h=1$ and $p=6$), whereas for a prediction 3 hours before $t=t_0$, one may use at most 4 measurements ($h=3$ and $p=4$), and so on.
Indeed, the number of lines, $p$, to keep in the secondary table depends on the value of $h$.

Notes: (i) using the same value for $p$ and $h$ (see Figure \ref{fig:schema}) for all patients does not in any way reduce the scope of the analysis.  The proposed approach works in exactly the same way if $p$ (with $h$ set to the same value for fairness to patients) were to vary from patient to patient (if the latter had more than 4 measurements in the past before $t_{-3}$). (ii) The proposed approach could also be multi-modal. Indeed, it would suffice to have one (or more) other table(s) containing alternative representations of patients and to apply the same process.
\vspace{-4mm}

\subsection{Flatening the data}

Machine learning models typically require input data in a tabular, flat format where each row represents an instance and each column represents a feature. Relational data stored across multiple tables needs to be combined into a single table to be used effectively for training classifiers. This action is also named propositionalization.
The propositionalization approaches consist of learning a model from relational data by flattening the original relational data, which are stored in several linked tables (similarly to databases)~\cite{lachiche10}. These approaches come from the field of Relational Data Mining~\cite{dvzeroski2009relational} and are not usually applied to time series. More precisely, relational data contains at least one root table where each row represents a statistical individual (e.g. `id` of the patients) and another secondary table containing detailed records (eg. the measurements realized of patients which are represented by each row of the secondary table). The specificity of relational data involves one-to-many relationships between the tables (eg. a patient underwent several identical medical examinations over time). The propositionalization problem consists of transforming the relational data into a single attribute-value dataset in order to use regular machine learning methods. 

Two kinds of propositionalization approaches can be distinguished (i) \emph{logic-based} methods, such as RSD~\cite{lavravc2002rsd}, SINUS~\cite{dzeroski1994inductive}, tackle the propositionalization task by constructing first-order logic attributes; (ii) the \emph{database-inspired} methods such as RELAGGS~\cite{krogel2001transformation} apply aggregation functions, such as Min, Max, and Mean in order to generate attributes. The interested reader can find a complete state-of-the-art and a comparative evaluation in~\cite{krogel2003comparative}. 

In our study, we used a Minimal Description Length (MDL) based propositionalization approach presented in~\cite{Boulle2018}. This approach exploits a Bayesian formalism to generate informative aggregate variables in a supervised way. To the best of our knowledge, this approach is the only one that avoids overfitting problems by regularizing the complexity of the generated variables. 

Note - Methodological clarification: We stress that the use of a Fractional Naive Bayes classifier should not be understood as ignoring complex dependencies among clinical variables. In our framework, a substantial part of these dependencies is already captured upstream by the MDL-based relational propositionalization step, which constructs conditional aggregate features encoding contextual, temporal, and non-linear relationships. Moreover, the Naive Bayes assumption is often informally overstated as a general independence assumption, whereas it is more precisely a class-conditional independence assumption. Hence, the assumption applies to the final constructed representation given the class, not to the raw clinical variables themselves. Of course, as with any explicit feature-based representation, some very high-order interactions or fine-grained temporal patterns may remain only partially represented if they are not reflected in the learned aggregates.

Our relational data (see Fig.~\ref{fig:schema}) consists of the root table, which contains $N$ instances, characterized by two variables: the patient identifiers and the target variable (i.e., class values). The secondary table contains $p N$ detailed medical examination, described by 36 variables already described in Section \ref{sec:protocol}. The number of aggregate variables to be constructed ($Q$) is the only user parameter. Our relational data is transformed into a regular attribute-value dataset by applying the MDL-based propositionalization approach. 
\vspace{-4mm}

\subsection{Variable selection and learned classifier}
\label{sec:vs}

The used MDL approach is able to select the most informative aggregate variables in two ways: i) by filtering uninformative aggregate variables; ii) by finding the most informative and independent subset of variables. 

\subsubsection{Filtering uninformative variables:} 
The filtering of the aggregate variables is based on previously developed supervised discretization~(\cite{BoulleML06}) and grouping~(\cite{BoulleMLDM05}) methods. This Bayesian approach turns the learning task into a model selection problem. A prior distribution is defined on the model space that exploits the hierarchy of its parameters. In practice, this approach reaches a good trade-off between robust and accurate models. The prior favors simple models with few intervals, and the likelihood favors models that fit the data regardless of their complexity. The aggregate variables are evaluated, one by one, using a specifically designed MDL optimization criterion~(\cite{Boulle2018}). The complexity of the aggregate variables is taken into account by adding a construction cost in the prior. This criterion can be interpreted as a coding length according to information theory. Compression Gain (CG) compares the coding length of the learned model with the empty model that includes a single interval. CG measures the ability of the learned models to compress the training data, despite the additional construction cost. Only the $R$ variables with a positive CG are retained ($R \leq Q$) 
 
\subsubsection{Finding the most informative subset of variables:}

All the $R$ informative variables coming from the previous step (after discretization or grouping values) are gathered together and used to learn a Fractional Naive Bayes classifier (a Naive Bayes which uses a subset of the $Q$ variables defined by a selective process). The Fractional Naive Bayes (FNB) aims to select the most informative and independent subset of variables by using a specifically designed MDL optimization criterion~(\cite{hue2024fractionalnaivebayesfnb}). It selects the most informative and independent subset of variables using a soft selection scheme with variable weights ranging between 0 and 1. The best model is obtained by optimizing a sparse regularization of the model's log-likelihood. The optimization algorithm employed consists of a sequence of forward and backward selection steps, which add or remove variable weights, starting with weight increments of 1. These two selection steps are repeated with decreasing weight increments, each time beginning with a random ordering of the variables. In the end, we retain the most probable subset of weighted variables that comply with the naive Bayes assumption, i.e., that are both informative and independent. At the end, we keep the most probable subset of variables compliant with the naive Bayes assumption, i.e. both informative and independent. This subset contains $S$ aggregate variables, with $S \leq R \leq Q$.

\subsubsection{Learning classifier:} \label{nb}
Finally, the used classifier is a naive Bayes classifier, which takes the $S $ selected aggregate variables as an input. As shown in Eq.~\ref{equation_nb}, the naive Bayes classifier~(\cite{Langley1992}) estimates the distribution of a particular class value $C_z$ conditionally to the input variables $x_k$. 

\begin{equation}
\fontsize{9}{9}\selectfont
	P(C_z|x_k)=\frac{P(C_z)\prod_{j=1}^{d} P(V_j=x_{jk}|C_z)^{W_j} }{\sum_{t=1}^{C} \left[ P(C_t)\prod_{j=1}^{d} P(V_j=x_{jk}|C_t)^{W_j} \right] }
	\label{equation_nb}
\end{equation}
$C$ is the number of class values to be predicted (in this paper $C=2$), $V$ the input variables, and $W$ the weights on the variable coming from the process described in \cite{hue2024fractionalnaivebayesfnb}. This simple and efficient classifier makes the assumption that the distributions $P(V_j=x_{jk}|C_z)$ are independent. In practice, these conditional distributions are estimated in a frequentist way, by using the previously learned univariate discretization~(\cite{BoulleML06}) and grouping~(\cite{BoulleMLDM05}) models. The denominator of Eq.~\ref{equation_nb} normalizes the estimated probability by making a sum of the numerator term over all the class values. At the end, the predicted class value given a particular $x_k$ is the one that maximizes the conditional probabilities $P(C_z|x_k)$. \\




\section{Results - Classifier performances}
\label{sec:results}

The average results of the classifier on a 10-fold cross-validation are presented in Table 1 
in terms of both accuracy and AUC, compared to the value of the number of $Q$ variables to be constructed (as described in the previous section). 
\begin{figure}[!h]
\centering
\begin{minipage}{.34\textwidth}
Obviously, the performances are very good. If a tradeoff between the number of variables incorporated in the model ($R$) and the AUC in test is required, then the values $Q$=1000, $R$=872 and  $R$=98 could be considered. 
\end{minipage}
\hfill
\begin{minipage}{.62\textwidth}
\centering
    \fontsize{7.5}{7}\selectfont
     \begin{tabular}{|c|c|c|c|c|c|c|}\hline
 Q	            &R	&S	&Acc Train 	&Auc train 	&Acc Test	&Auc Test \\ \hline
10	    &5	     &5	    &0,7941	&0,8767	&0,7719	&0,8526 \\ \hline
100	    &86	     &44	&0,9379	&0,9832	&0,8963	&0,9664 \\ \hline
1000	&872	 &98	&0,9620	&0,9918	&0,9009	&0,9708 \\ \hline
10000	&8849	 &198	&0,9713	&0,9969	&0,9171	&0,9751 \\ \hline
100000	&72331	 &272	&0,9883	&0,9992	&0,9194	&0,9737 \\ \hline
     \end{tabular}
     \captionof{table}{AUC Results versus the number of Q variables to be constructed.}
     \label{tab:results}
\end{minipage}
\end{figure}
The aggregated variables (98) to be constructed are sufficient to obtain good performance and a frugal model. But it is up to the final user since the largest model has only 272 input variables. The section \ref{aggregates} will present some of these aggregate variables.\\

\subsection{Comparative Evaluation Against Different Models: }

Recent advances in machine learning have led to several commercial systems for early prediction of sepsis, evaluated in heterogeneous clinical settings. Among the most representative are \textit{InSight\textsuperscript{\textregistered}} \cite{desautels}, \textit{Sepsis ImmunoScore\texttrademark} \cite{Taneja2021Sepsis}, \textit{NAVOY\textsuperscript{\textregistered}} Sepsis \cite{Persson2024NAVOY} and \textit{VFusion\texttrademark} Sepsis \cite{VFusionSepsis} report high performance, with AUC values generally between 0.83 and 0.91. 
Highlighted by the PREVIA \cite{thiboud2025development} retrospective study and in line with existing meta-analyses \cite{Fleuren2020Meta}, direct comparison of these systems remains challenging due to substantial differences in patient populations, prediction horizons, variables used, and validation protocols. 

Besides these commercial systems, the Table~\ref{tab:comparison} summarizes the performance of the proposed 
approach against four competitive baselines on the same experimental 
setup (10-fold cross-validation, MIMIC-III, Q=1000 aggregate variables 
for feature-based methods). Several observations emerge. First, the proposed MDL-FNB approach 
achieves an AUC of 0.983 on the test set, which is competitive with 
CatBoost (0.985) and XGBoost (0.985), and markedly superior to LSTM 
(0.945). Second, while tree-based ensembles (RF, XGBoost, CatBoost) 
reach a perfect training AUC of 1.0, their substantially higher 
train-test gap reveals a tendency toward overfitting, whereas Khiops 
maintains a train AUC of 0.991  indicating better generalization. 
Third, the proposed model achieves this performance with only 98 
selected variables and a serialized model size of 1.0 MB, compared to 
996 variables and 10.2 MB for the best Random Forest configuration. 
Finally, robustness scores (Rob. AUC) are consistently high across 
methods, with Khiops matching the top score of 0.99. 

The size of the Neural Netwwork (NN) is 1000-256-2, this size and the number of layers have been obtained using cross-validation. The activations are Linear+BatchNorm1d and Linear for the last one. The loss function is the CrossEntropyLoss. The size of LSTM (H=64) has also been obtained using cross-validation. All the used codes are available in the \href{https://anonymous.4open.science/r/Sepsis-2026-7E45/README.md}{GitHub}.

\begin{table}[!h]
\centering
\fontsize{7}{8}\selectfont
\begin{tabular}{|l|c|c|c|c|c|c|c|}
\hline
\textbf{Model} & \textbf{\#Var} & \textbf{AUC Train} & \textbf{AUC Test} 
  & \textbf{Rob} & \textbf{Size} & \textbf{I} &  \textbf{E} \\
\hline
LSTM (H=64)          & 34   & 0.972 & 0.945 & 0.97 & 26 (pth)  & \xmark  & 1.027 Wh\\
Random Forest        & 996  & 1.000 & 0.982 & 0.98 & 3571 (pickle) & partial (2) & 0.173 Wh\\
XGBoost              & 607  & 1.000 & 0.985 & 0.98 & 268 (json)  & partial (2) & 0.135 Wh\\
CatBoost             & 860  & 1.000 & 0.985 & 0.98 & 1183 (cbm) & partial (2) & 1.503 Wh\\
TabPFN-2.6           & 1000 & 1.000 & 0.988 & 0.98 & 354.41 (pickle) & ? & 1444.926 Wh\\
Neural Network   	 & 1000	& 1.000 & 0.934	& 0.93 & 2 (pth) & partial (2) & 1.114 Wh \\ \hline
\hline
\textbf{Khiops (ours)} & \textbf{98} & \textbf{0.991} & \textbf{0.983} 
  & \textbf{0.99} & \textbf{234} (ascii) & \textbf{\cmark (4 levels)} & \textbf{0.150 Wh}\\
\hline
\end{tabular}
\captionof{table}{Comparison of models at Q=1000 - Legend: (i) Rob: $\mbox{AUC}_{\mbox{Test}}$ / $\mbox{AUC}_{\mbox{Train}}$. (ii) Size: model size in ko. (ii) I: Partial interpretability refers 
to post-hoc methods (e.g., SHAP); \cmark~denotes native, intrinsic interpretability. (iv) E: Energy consumed to train the model (library code carbon). It should be noted that the value for the energy are without the grid search for the NN and all models (except the LTSM) benefit of the aggregates provided by Khiops. For Khiops, the value includes the aggregate construction.}
\label{tab:comparison}
\end{table}

Crucially, as the following sections will show,  while gradient boosting and random forest methods offer only partial, post-hoc interpretability through tools such as SHAP 
\cite{lundberg2017}, our approach provides four intrinsic, 
model-native levels of explanation: univariate, global, local, and 
counterfactual without any approximation. This combination of competitive predictive accuracy, 
model compactness and full interpretability positions the proposed 
approach as particularly suited for clinical deployment.

The propositionalization step intentionally transforms longitudinal trajectories into an explicit aggregate representation, The MDL-based construction process is supervised and designed to retain aggregates that are both informative and parsimonious, including conditional statistics such as extrema, variability, and contextualized summaries. In practice, these constructed features capture an important part of the temporal signal relevant for prediction, as suggested by the competitive performance observed against, for example, the LSTM.\\

\subsection{Discussion of model parsimony}

The models reported in Table \ref{tab:comparison} should not be viewed as simple baselines in the weak sense of the term. Rather, they constitute strong downstream comparison partners spanning several competitive learning families, including feed-forward neural networks, random forests, gradient-boosted trees and khiops. Moreover, these classifiers were evaluated under the same experimental protocol and, for the feature-based models, on the same MDL-based propositionalized representation, which allows a meaningful comparison of downstream predictive behavior while controlling for data preparation. In particular, CatBoost and XGBoost are widely recognized as high-performance learners on structured clinical data, and their inclusion provides a demanding benchmark rather than a minimal reference point. The fact that the proposed MDL-FNB pipeline remains competitive with these strong models, while using far fewer variables and offering native interpretability, should therefore be interpreted as a substantive result rather than as an improvement over weak baselines.

At the same time, we acknowledge that the comparative evaluation could be further extended to include more specialized sequence-based deep learning approaches for irregular clinical time series, such as MGP-RNN~\cite{futoma2017} and MGP-TCN~\cite{moor2019early}. These methods were specifically designed for early sepsis prediction from temporally irregular observations and therefore provide a useful complementary benchmark with respect to end-to-end temporal representation learning. In future work, we plan to broaden the comparative study by evaluating such architectures under a harmonized experimental setting, using the same cohort construction, labeling procedure, prediction horizons, and evaluation metrics. This extension would strengthen the positioning of the proposed framework relative to the temporal deep-learning literature, while also helping to clarify the respective roles of MDL-based feature construction and downstream classification.

\section{Results - Interpretability}

\subsection{Example of aggregates elaborated}
\label{aggregates}
The Table \ref{tab:importance}, column 1, gives the $S$ 20 most important aggregated variables found by the approach  used~\cite{Boulle2018} here. A nice property is their straightforward meaning.

\begin{figure}
\centering
\begin{minipage}{.34\textwidth}
 On this dataset they only have two operands, even if the method in \cite{Boulle2018} may produce more complex ``rules''. Each of them is easy to read and to understand even for a non-data scientist (or research scientist) and moreover, in terms readable by medical practitioners. We also give for the first one a color code where the blue parts are the rules and the orange parts the variables present in the secondary table (to ease the reading). It is worth noting that the clinical interpretation of some of these rules still needs to be validated using longitudinal data collected between 2022 and 2025 at XXXX\footnote{Hospital name hidden for anonymity reasons during submission.} Hospital.
\end{minipage}
\hfill
\begin{minipage}{.62\textwidth}
    \centering
    \fontsize{6}{7}\selectfont
    \begin{tabular}{l|c|c}\hline
Name	&Level	&Imp \\ \hline
\textcolor{blue}{Min}(\textcolor{orange}{EtCO2}) \textcolor{blue}{where} \textcolor{orange}{Calcium} \textcolor{blue}{$>$ 8.075} & 0.1999	&0.02909\\ 
Min(HospAdmTime) where BaseExcess $>$ -1.61	&0.0976	&0.01741\\ 
Min(Temp) where Hct $\leq$ 29.505	&0.1275	&0.01671\\ 
StdDev(EtCO2) where SaO2 $\leq$ 95.79	&0.1490	&0.01601\\ 
Max(O2Sat) where HR $\leq$ 86.95	&0.1412	&0.01572\\ 
StdDev(O2Sat)	&0.07872	&0.01492\\ 
Max(EtCO2)	&0.1650	&0.01467\\ 
StdDev(FiO2) where DBP $>$ 59.95	&0.1023	&0.01458\\ 
Max(FiO2) where O2Sat $>$ 99.05	&0.3774	&0.01451\\ 
Min(Temp) where Age $>$ 65.495	&0.1214	&0.01428\\ 
StdDev(BUN) where SaO2 $>$ 95.79	&0.01601	&0.01331\\ 
StdDev(FiO2) where Fibrinogen $\leq$ 229.15	&0.1424	&0.01252\\ 
Min(pH) where Lactate $>$ 2.496	&0.06828	&0.01216\\ 
StdDev(Alkalinephos) where Glucose $>$ 131.95	&0.02323	&0.01199\\ 
Mean(MAP) where SaO2 $>$ 95.79	&0.02699	&0.01155\\ 
Min(Potassium) where Chloride $\leq$ 108.55	&0.02291	&0.01134\\ 
Sum(Calcium) where BUN $>$ 19.3	&0.04739	&0.01127\\ 
StdDev(FiO2) where Calcium $>$ 8.075	&0.1552	&0.01103\\ 
Sum(Temp) where AST $\leq$ 167.5	&0.1131	&0.01093\\ 
Min(pH) where Potassium $\leq$ 4.162	&0.09446	&0.01087\\  \hline
    \end{tabular}
    \captionof{table}{The $S$ 20 most important aggregated variables}
    \label{tab:importance}
\end{minipage}
\end{figure}


\subsection{Univariate variable importance}

\begin{figure*}[!h]
    \centering
    \includegraphics[width=0.95\linewidth]{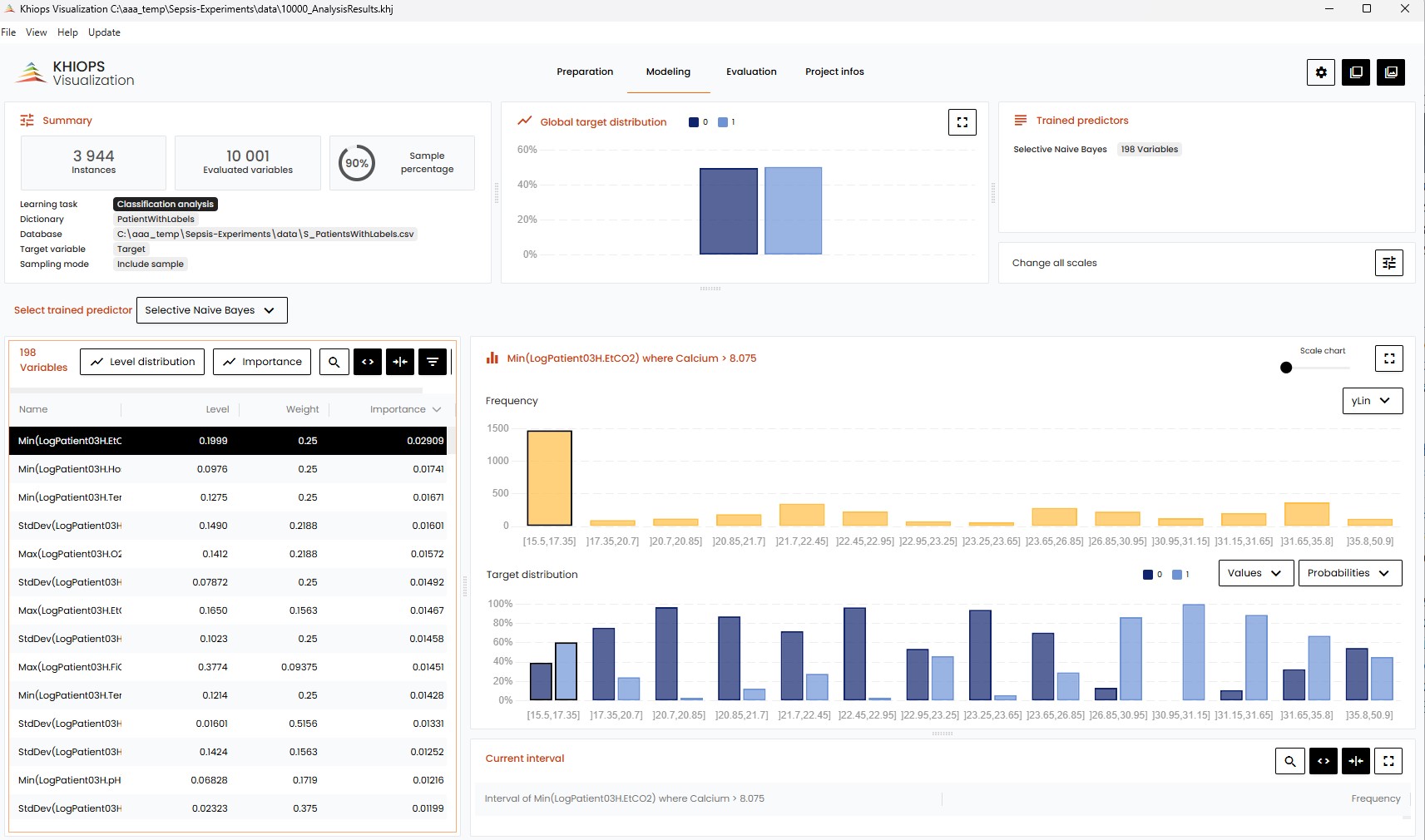}
    \caption{Screenshot of Khiops Visualization after analyzing the sepsis database and constructing 10000 aggregates}
    \label{fig:visu}
\end{figure*}

As described in Section \ref{sec:vs} the Khiops classifier is trained in two steps. The first one is to evaluate the univariate predictive information contained in each aggregated variable. For that, a supervised discretization~\cite{BoulleML06} is realized for numerical one and a supervised grouping  method~\cite{BoulleMLDM05} is performed on categorical variables. In the dataset used here all the variables are numerical (except the 'Gender' and `Hour' variables redrawn at the beginning of the study). One output of this preprocessing is an information of the univariate predictive information, named ``level'', of each variable, which is in the range $[0:1]$ (0 variable no informative, 1 variable 100 \% informative). These values are given in the second column of Table 1. 
Since each aggregated variable is discretized, it is also possible to have a look at the information carried by the intervals found. For that, the Khiops library provides an interactive results visualization tool, called Khiops Visualization. We present in Figure \ref{fig:visu} a copy of a screenshot of this tool for the most important aggregate: $V_1$=``Min(EtCO2) where Calcium $>$ 8.075''.
The analysis of the discretization allows us to understand why this aggregate is informative. Small values of $V_1$ produce intervals relatively pure of Sepsis=1 while the intervals between 26 and 36  are relatively pure of Sepsis=0. The reading of $P(X|C)$ per interval, and for each variable, is informative to understand the result of the univariate classifications.

\subsection{Local variable importance}

To analyze local behavior (for each example), it is possible to compute the Shapley values of all the input variables of the trained classifier using the method described in \cite{10.1007/978-3-031-74630-7_6} and available in the Khiops library V11.
This allows to have the variable importance example by example (local importance).
We illustrate this information (XAI) in the Table \ref{tab:localshapley}.

\begin{table*}[!ht]
    \centering
\resizebox{\linewidth}{!}{
    \begin{tabular}{|c|l|c|c|l|c|c|}\hline
ProbTarget1	& ShapleyVariable\_1 &ShapleyPart\_1	&ShapleyValue\_1	&ShapleyVariable\_2	&ShapleyPart\_2	&ShapleyValue\_2 \\ \hline
0,999936	&Min(EtCO2) where Calcium $>$ 8,075	&]30,95,31,15]	&3,2213 &Min(DBP) where DBP $>$ 59,95	&]66,55,66,95]	&1,9981\\ \hline
0,999936	&Max(EtCO2)	&]31,35,31,55]	&0,5986 &Max(EtCO2) where MAP $>$ 77,16	&]31,35,31,55]	&0,5727\\ \hline
0,999929	&Median(Bilirubin\_direct) where HospAdmTime $>$ -7,64	&]4,27,4,33]	&1,4618	&Min(HCO3) where Glucose $>$ 131,95	&]22,015,22,195]	&1,2114\\ \hline
0,999929	&Min(EtCO2) where Calcium $>$ 8,075	&]30,95,31,15]	&3,221	&Min(Temp) where Age $>$ 65,495	&]37,283,+inf[	&0,6948\\ \hline
0,999929	&Min(EtCO2) where Calcium $>$ 8,075	&]30,95,31,15]	& 3,221	&Max(O2Sat) where HR $\leq$ 86,95	&]-inf,98,1]	&0,4083\\ \hline
    \end{tabular}}
    \captionof{table}{Illustration of the knowledge provided by the local importance}
    \label{tab:localshapley}
\end{table*}

Five examples among those assigned a high predicted probability of sepsis are presented. 
The first column gives the predicted probability of the sepsis-positive class. Then, for each patient, the two variables that contribute the most to the predicted probability (the number of variables is just defined per user when asking this XAI outcome in the library (see the code in the GitHub provided for reproducibility)). Therefore, here, after the first column, there are 2 triplets of columns. Each triplet gives for each patient the name of the variable, then the value of the variable and finally the Shapley value for this variable. The triplets (so the columns of the file) are sorting according to the Shapley value, allowing a fast understanding of the individual variable importance.

When examining, for example, the second patient (line 1) in this table, we see that the most important variable is ``Min(EtCO2) where Calcium $>$ 8,075 '' and the second one is ``Min(DBP) where DBP $>$ 59,95 ''. The value of the most important belongs to the value interval ]30,95,31,15]  while the value of the second most important value belongs to the value interval ]66,55,66,95] . The associated Shapley values are in columns 4 and 7. For this patient, the main contributors to the high predicted probability of sepsis are therefore very easy to understand. The other lines of this table appear to be equally straightforward to read.

Note: the Khiops library can also output a file with all the Shapley values for all variables and for all the classes, allowing the use of this file with a Python library like Shap \cite{lundberg2017} to create personalized visualization.

\subsection{Global variable importance}

The global importance of the variables is given in column 3  of the Table \ref{tab:importance}. These importance are defined for each variable as the average on all the train samples of the absolute values of the local Shapley values described in the previous section.

\subsection{Counterfactuals}

Counterfactual reasoning is a concept from psychology and social sciences \cite{xaisocial}, which involves examining possible alternatives to past events \cite{stepin2021survey}. Humans often use counterfactual reasoning by imagining what would happen if an event had not occurred, and this is precisely what counterfactual reasoning is. When applied to artificial intelligence, the question is, for example, ``Why did the model make this decision instead of another?" (counterfactual explanation) or ``How would the decision have differed if a certain condition had been changed?". This reasoning can take the form of a counterfactual or semi-factual explanation.

A counterfactual explanation might be ``If your income had been \$10000 higher, then your credit would have been accepted'' 
\cite{LemaireIGI2010correlation}.A semi-factual is a special-case of the counterfactual in that it conveys possibilities that ``counteract'' what actually happened, even if the outcome does not change \cite{aryal2023explanations}: ``Even if your income had been \$5000 higher, your credit would still be denied'' (but closer to being accepted). 
Within the framework of counterfactual reasoning, we here used the method described in \cite{Lemaire2023ViewingTP} and the notion of trajectory. Indeed, the study of counterfactual trajectories a posteriori is of great interest, as it also makes it possible to identify when a trajectory is approaching the frontier or pass trough the frontier.
In this paper, the trajectory of a counterfactual is the minimum of changes in the input vector to change the predicted class to the opposite\footnote{A notebook that performs this process for the Khiops library is given in the Github for reproducibility}. 
In the case of Sepsis, one could be interested, for example, when examining Patients :
\begin{itemize} 
\item predicted as negative (Target=0) but close to the decision boundary: what are their counterfactuals leading to Target=1? In other words, which changes would be sufficient to reverse the prediction?
\item predicted as positive (Target=1), for whom one may wish to identify changes that would reverse the prediction to Target=0.
\end{itemize}
We give below one example of these two possibilities when one considers all the input variables (of the input vector) as ``alterable''. But as explained in the literature and in \cite{Lemaire2023ViewingTP} it is up to the user to limit the list of the alterable variables since some of them could not be changed. This must be done in consultation with the expert of the application domain, here a patrician.\\

\noindent In the first case :
{\small
\begin{itemize}
    \item for the patient `10001' (Patient\_Id) : Initial class = '0' - Proba '0' = 0.561962
    \item the trajectory for the counterfactual as a single stage : 
    \begin{itemize}
    \item Step 1 : if  ``StdDev(EtCO2)'' change of value from 5.795418449 to 0.2537
    \end{itemize}
\end{itemize}}

After the first stage, the initial predicted probability of the negative class (0.561962) decreases to 0.131336. The predicted class therefore switches to sepsis-positive.\\

\noindent In the second case
{\small
\begin{itemize}
    \item for the patient `100003' (Patient\_Id) : Initial class = '1' - Probability '1' = 0.999787
    \item the trajectory for the counterfactual as 6 stages :
    \begin{itemize}
\item Step 1 : `Min(Temp)' from 37.6 to 31.3135 $\rightarrow$ Proba '1' = 0.999021, 
\item Step 2 : `Max(HospAdmTime)' from -253.56 to 8.022499 $\rightarrow$ Proba '1' = 0.994351, 
\item Step 3 : `Min(pH)' from 7.398 to 7.269   $\rightarrow$ Proba '1' = 0.972920, 
\item Step 4 : `StdDev(EtCO2)' from 2.965952629 to '8.17991062'  $\rightarrow$ Proba '1' = 0.912302, 
\item Step 5 : `Median(FiO2)' from 0.4 to 1.0095  $\rightarrow$ Proba '1' = 0.751472, 
\item Step 6 : `Median(HospAdmTime)' from 253.56 to 8.022499  $\rightarrow$ Proba '1' = 0.471857, 
    \end{itemize}
After the 6th stage, the initial predicted probability of the positive class decreases to 0.471857. The predicted class therefore switches to sepsis-negative.\\
\end{itemize}
    }

The information provided by the counterfactual trajectory is useful for understanding which variable changes would be sufficient to modify the model's sepsis prediction.

\subsection{Summary on interpretability}

Overall, the previous analyses illustrate that the proposed framework provides four complementary and model-native levels of interpretability: univariate, global, local, and counterfactual. This point is important when positioning our work with respect to recent sepsis prediction studies. For example, Hu et al.~\cite{hu2022interpretable} developed an XGBoost-based prognosis model and analyzed it using SHAP, while Zhou et al.~\cite{zhou2025early} reported externally validated septic shock prediction together with SHAP-based explainability. These contributions clearly improve the transparency of high-performing machine-learning models. However, they mainly provide post-hoc global and instance-level explanations of complex predictors, whereas our framework is interpretable by construction and explicitly integrates four complementary levels of explanation within the same modeling pipeline. Therefore, our claim is not that existing deep or boosted models are entirely non-interpretable, but rather that their explainability is of a different nature from an intrinsic, multi-level, explainable-by-design approach.

\section{Conclusion}
\label{sec:conclusion}
In this study, we presented an innovative approach for the early detection of sepsis by exploiting electronic medical data in relational form. By combining the relational representation of patient trajectories, the propositionalization technique based on the MDL method, and a selective Bayesian classifier, we succeeded in obtaining a model that is both effective and intrinsically interpretable. This approach not only improves prediction accuracy but also provides a clear understanding of the key factors contributing to sepsis detection, thereby facilitating acceptance by clinicians.
Experimental results demonstrate that the proposed approach matches 
the predictive accuracy of state-of-the-art black-box models while 
remaining fully interpretable, compact, and robust  three properties 
that are rarely achieved simultaneously in clinical machine learning.
In addition, the flexibility of relational representation opens up prospects for the integration of multimodal data and the management of patient trajectories of varying sizes.

The proposed approach is based on a four-dimensional interpretation: univariate, global, and local. The first step, univariate, consists of analyzing the importance of each individual variable by evaluating their predictive contribution using measures such as univariate information. The second step, global, provides an overview of the most influential variables across all patients. Finally, local interpretation focuses on each patient individually, using methods such as Shapley value analysis to identify the specific variables that contributed most to the model's decision for that particular case. This three-step approach provides a comprehensive and hierarchical understanding of the model's mechanisms, facilitating its acceptance by clinicians and its implementation in a medical context.

In the context of sepsis, counterfactuals offer the possibility of identifying the changes needed in clinical variables to shift a prediction from positive to negative (or vice versa). For example, by adjusting certain physiological parameters of a patient, it becomes possible to understand which factors most influence the model's decision. This approach promotes a better understanding of the underlying mechanisms and can guide medical interventions by proposing alternative trajectories to improve care.
This study, which incorporates counterfactual reasoning and local explanation techniques, helps to increase transparency and confidence in predictive models in clinical settings. 

Several limitations of this study should be acknowledged. First, the cohort of 3940 patients was obtained through a strict selection criterion on temporal completeness, which may limit generalizability to patients with sparser clinical records. This strict temporal completeness criterion was adopted to ensure label reliability and to make horizon-specific comparisons methodologically consistent across patients. We acknowledge that such regular sampling is not always observed in routine clinical practice, where measurement frequency may be irregular, sparse, or delayed depending on patient status and care organization. As a result, the present cohort should be viewed as a controlled evaluation setting, and performance may differ when the method is applied to less complete real-world trajectories or to datasets with substantially different class prevalence, which may also affect probability calibration. The methodology here presented will be applied, in future works, on irregular and  sparse trajectories of patients.

Second, all experiments were conducted on a single dataset (MIMIC-III), and external validation on independent cohorts remains necessary to assess the robustness of the approach across different hospital settings and patient populations. While one of the authors is a clinician specializing in sepsis who validated the clinical relevance of the identified features, a prospective clinical evaluation of the full decision-support pipeline falls outside the scope of this work but remains an important direction for future work.

Future work should also investigate the impact of missing-data handling choices. In the present study, variables with more than 20\% missing values were discarded and remaining missing observations were imputed using nearest-neighbor interpolation, yielding a pragmatic and homogeneous preprocessing pipeline. However, we did not assess whether alternative exclusion thresholds or more advanced imputation strategies could improve predictive performance or preserve additional clinically relevant information. Since missingness patterns in Electronic Health Record data may themselves be associated with covariates, care pathways, and sepsis status, future studies should examine these mechanisms explicitly and evaluate preprocessing strategies that account for informative missingness.

{\footnotesize {\bf Note on reproducibility: } The initial dataset (MIMIC-III), which contains 40000 patients, as well as the 3940 patients used in the experiments, is available on \href{https://anonymous.4open.science/r/Sepsis-2026-7E45/README.md}{GitHub}. The code to train the classifiers (Table \ref{tab:results}), output the local importance (Table \ref{tab:localshapley}), the counterfactuals, etc,  is also available in the same GitHub. Thus, all results are fully reproducible. Moreover, one of the authors of this paper, Pierre Jaquet, is a doctor specialist in Sepsis who validated the findings.}

\bibliographystyle{splncs04}

\bibliography{refs}

\end{document}